\pdfoutput=1

\documentclass[11pt]{article}

\usepackage{emnlp2021}

\usepackage{times}
\usepackage{latexsym}

\usepackage[T1]{fontenc}

\usepackage[utf8]{inputenc}

\usepackage{microtype}
\usepackage{multirow}
\usepackage{graphicx}

\title{Modern Question Answering Datasets and Benchmarks: A Survey}

\author{Zhen Wang\\
  Delft University of Technology \\
  \texttt{z.wang-42@student.tudelft.nl}}

\begin{document}
\maketitle
\begin{abstract}

Question Answering (QA) is one of the most important natural language processing (NLP) tasks. It aims using NLP technologies to generate a corresponding answer to a given question based on the massive unstructured corpus. With the development of deep learning, more and more challenging QA datasets are being proposed, and lots of new methods for solving them are also emerging. In this paper, we investigate influential QA datasets that have been released in the era of deep learning. Specifically, we begin with introducing two of the most common QA tasks - textual question answer and visual question answering - separately, covering the most representative datasets, and then give some current challenges of QA research.

\end{abstract}

\section{Introduction}

Question answering (QA) \cite{qa} aims at providing correct answers to questions base on some given context or knowledge. QA is a traditional research direction that has been proposed half a century ago. People hope to help with everyday life by teaching the program how to answer questions like a real person. Traditional QA systems integrate some information retrieval techniques to find answers. With the development of deep learning, computer programs can now tackle more complex problems. At the same time, in the era of deep learning, more and more datasets are being proposed to measure the capabilities of QA models. These datasets, in turn, facilitate the development of deep learning QA models. 

An comprehensive understanding of those datasets and benchmarks is essential before the further research about QA. Therefore, in this paper, we investigate some of the most commonly used datasets nowadays and categorize them according to the capabilities of the QA models involved. Meanwhile, according to the different modes designed, we divide them into three categories: textual QA, image QA, and video QA.
In textual QA, all the corpora involved are presented in textual form. A typical sample in textual QA consists a question, an answer, and a paragraph that contains the answer. Image QA and video QA together are generally referred to as Visual Question Answering (VQA) \cite{vqa}. In image QA, the question and answer are usually in textual form while the context is an image. And in video QA, the question and answer are the same but the context is a clip of video.
We will start by introducing some representative datasets for each of the three types of QA, and conclude by analyzing some of the current challenges and opportunities in QA research from a dataset perspective.
We hope this paper will help researchers gain a comprehensive understanding of QA tasks and methods, attracting more attention, leading to greater progress in QA filed.

\section{Textual Question Answering}

\begin{table*}[ht]
\centering
\small
\begin{tabular}{lllll}
\hline
\textbf{Dataset} & \textbf{Answer Type} & \textbf{Size} & \textbf{Domain} & \textbf{Evaluate Ability} \\ \hline
ARC\cite{arc} & Multi-Choice & 7,787 & Science & Reasoning \\
BoolQ \cite{boolq} & Bool & 16K & Wikipedia & Reasoning \\
BioASQ \cite{bioasq} & Span & 282 & Biomedical & Articles Indexing \\
CaseHOLD \cite{casehold} & Multi-Choice & 53,137 & Law & Pre-training \\
bABi \cite{babi} & Bool/Entity & 40K & Open Domain & Reasoning \\
CBT \cite{childrensbook} & Entity & 20K & Children's Book & Model Memory\\
CliCR \cite{clicr} & Entity & 105K & Medical & Domain Knowledge \\
CNN and Daily Mail \cite{cnndaily} & Entity & 311K & News & Text Summarization \\
CODAH \cite{codah} & Multi-choice & 4,149 & Open Domain & Commonsense\\
CommonsenseQA \cite{commonsenseqa} & Multi-choice & 12,247 & ConceptNet & Commonsense\\
\begin{tabular}[c]{@{}l@{}} ComplexWebQuestions \\ \cite{complexwebquestions} \end{tabular} & Entity & 34,689 & Freebase & Multi-hop\\
ConditionalQA \cite{conditionalqa} & Entity/Span & 9983 & Public Policy & Multi-hop \\
COPA \cite{copa} & Multi-choice & 1000 & Commonsense & Reasoning \\
CoQA \cite{coqa} & Entity & 127K & Open Domain & Conversation\\
DROP \cite{drop} & Span & 96K & Wikipedia & Multi-hop\\
FinQA \cite{finqa} & Number/Span & 8,281 & Finance & Multi-hop\\
HotpotQA \cite{hotpotqa} & Entity & 113K & Wikipedia & Multi-hop\\
JD Production QA \cite{productqa} & Generation & 469,953 & E-commerce & Domain Knowledge \\
LogiQA \cite{logiqa} & Multi-choice & 8,678 & Exam & Reasoning \\
MCTest \cite{mctest} & Multi-choice & 2,000 & Fictional Story & Reading Comprehension \\ 
\begin{tabular}[c]{@{}l@{}} Mathematics Dataset \\ \cite{methematics} \end{tabular} & Numeric & $2.1 \times 10^6$ & Mathematics & Calculate \\
MS MARCO \cite{msmarco} & Generation & 1,010,916 & Web pages & Search \\
MultiRC \cite{multirc} & Multi-choice & 6K & Multiple Domain & Multi-hop \\
NarrativeQA \cite{narrativeqa} & Span & 46,765 & Story & Full Document \\
\begin{tabular}[c]{@{}l@{}} Natural Questions \\ \cite{naturalquestions} \end{tabular} & Span/Passage & 323,045 & Wikipedia & Search \\
NewsQA \cite{newsqa} & Span & 100,000 & CNN news & Reading Comprehension \\
OpenBookQA \cite{openbookqa} & Multi-choice & 6000 & Science Facts & Reasoning \\ 
PIQA \cite{piqa} & Multi-choice & 21,000 & Physical & Physical \\
PubMedQA \cite{pubmedqa} & Multi-choice & 1K & Medical & Summarization \\
QASPER \cite{qasper} & Extractive & 5,049 & NLP papers & Reasoning \\
QuAC \cite{quac} & \begin{tabular}[c]{@{}l@{}} Multi-choice \\ Generation \end{tabular} & 100K & Wikipedia & Dialog \\
QUASAR \cite{quasart} & Span & 43,000 & StackOverflow/Trivia & search \\
RACE \cite{race} & Multi-choice & 100,000 & Exam & Reading Comprehension \\
ReClor \cite{reclor} & Multi-choice & 6138 & Exam & Logical \\
SCDE \cite{scde} & Exam & 6K & Exam & Reading Comprehension \\
SimpleQuestions \cite{simplequestions} & Entity & 100K & Freebase & Knowledge \\
SQuAD \cite{squad1, squad2} & Span & 130,319 & Wikipedia & Reading Comprehension \\
TriviaQA \cite{triviaqa} & Span & 650K & Open Domain & Reading Comprehension \\
TweetQA \cite{tweetqa} & Generation & 13,757 & Tweet & Reading Comprehension \\
WikiHop \cite{wikihop} & Multi-choice & 51,318 & Wikipedia & Multi-hop \\
WikiQA \cite{wikiqa} & Sentence & 3,047 & Wikipedia & Reading Comprehension \\
\hline
\end{tabular}
\caption{Statistics of textual QA datasets.}
\label{table:qadataset}
\end{table*}

\subsection{Reading Comprehension}

BoolQ \cite{boolq} is a yes/no reading comprehension QA dataset. Give the passage ``The Great Storm of 1987 was a violent extratropical cyclone which caused casualties in England, France and the Channel Islands . . .'' and corresponding question ``Has the UK been hit by a hurricane?'', the correct answer is ``Yes''. A method is asked to find the correct answer though analyse complex and non-factoid textual information, thus strong inference ability is required. 

CNN and Daily Mail \cite{cnndaily} using human generated abstractive as questions while one entities is hided, the stories are used as context to do the fill-in-the-blank questions.

CoQA \cite{coqa} is the first conversational question answering dataset. Given a passage as the conversation context, one person will ask some questions and one person will answer those questions by find the evidences from context. Different from previous QA datasets, in CoQA, each question is related to previous asked questions, thus the answer should not only the context, but also the previous answered QA-pair should be taken into consideration. For example, the first question is `` Who had a birthday?'', and the second question is ``How old would she be?'', which one the ``she'' refer to can only be known by the first question.

MCTest \cite{mctest} is a multi-choice reading comprehension QA dataset. All the passages in MCTest are children fictional stories written by Amazon Mechanical Turk (AMT). For each story, four multi-choice questions are asked. Specifically, some questions are required must consider multiple sentences to get correct answers.

The candidate passages in MS MARCO \cite{msmarco} are all from the web pages retrieved by Bing, the questions are the query searched by Bing users, and the answers are generate by human annotators. There are three types of tasks in MS MARCO. First is determine whether a question is answerable, second is generate a proper answer to a give question, and the last is ranking the retrieved passages according to the relevance.

Given previous QA datasets can be answered with some QA irrelevant features such as term frequency, making it impossible to measure the true capabilities of QA models, Deepmind thus propose a new reading comprehension QA dataset called NarrativeQA \cite{narrativeqa}, where the questions can only be answered after fully review the whole document. 

Similar to MS MARCO \cite{msmarco}, Natural Questions (NQ) \cite{naturalquestions} is also a reading comprehension dataset collect through the search engine. And the major difference is that NQ not only provide the relevant paragraph as the long answer, but also a human annotated short answer. And the Wikipedia page is used as the context.

NewsQA \cite{newsqa} is aiming at providing a SQuAD-like QA dataset but harder than SQuAD \cite{squad1}. The annotators are asked to generate questions based on CNN news, and the answers are span of text which maybe the person, location, clause, numeric, etc.

RACE \cite{race} is a multi-choice reading comprehension dataset which collected from English exams for Chinese students. Based on a passage, models should find correct answer from four candidate answers to a given question.

Similar to RACE \cite{race}, ReClor \cite{reclor} is also a Multi-choice reading comprehension dataset. While some researches argue that current human annotated QA datasets contain some bias which may be used by models to cheat to achieve high accuracy, Reclor thus is divided into two datasets based on bias. The EASY set contains bias while the HARD set not. And the experiments prove that models can achieve high performance in EASY set but poor in HARD set.

SCDE \cite{scde} is used to test models' reading comprehension ability through fill up blanks in passage using given sentences. Given seven sentences where two sentences among them are distractors, models should select the correct five sentences and fill them in the corresponding blanks to complete the whole article.

SimpleQuestions \cite{simplequestions} is constructed using Freebase. For each (subject, relationship, object), two of them is used to generate the question and the remained one is used as the answer. The models need to retrieve the correct answers from massive of possible alternatives from freebase, which makes the questions are not easy to be answered.

SQuAD \cite{squad1, squad2} is the most famous reading comprehension dataset. Given a question and a paragraph as the context, the models need to extract the plausible answer from it. Each answer is a span of text. To make the task harder, in SQuAD 2.0, some unanswerable questions are include into the dataset.

TriviaQA \cite{triviaqa} is a reading comprehension dataset focusing on complex and compositional questions where requiring the ability on reasoning over multiple sentences. Specially, there are syntactic and lexical variability among questions, answers and evidences.

The question-answer pairs in TweetQA \cite{tweetqa} are all annotated from tweets. Unlike previous reading comprehension datasets using a text span as the answer, the answer in TweetQA can be abstractive, which makes it more difficult to be solve. The experiments show that current SOTA models such as BERT \cite{bert} are significantly below human performance. 

\subsection{Reasoning}

AI2 Reasoning Challenge (ARC) \cite{arc} focuses on hard multi-choice questions to challenge current QA methods. Their questions cannot be answered easily using retrieval based methods or through word correlation. And particularly, they choose some wrong answer question construct a specific Challenge Set contains 2590 questions.

CODAH \cite{codah} is a common sense reasoning dataset. Given a premise, for example, ``A man on his first date wanted to break the ice. He ...'', methods are asked to choose a correct reason from four answers. In this sceanario, the answer is ``made a corny joke.''

Combine totally 20 kinds of tasks, bABi \cite{babi} aiming at give the answer that can a QA model is able to solve QA problem using some reasoning abilities, such as the chain of facts, induction, deduction operation. When give three fact: (1) Mary went to the bathroom. (2) John moved to the hallway. (3) Mary travelled to the office; and the question ``Where is Mary?'', the tested QA models should give the correct answer ``office''.

Build through ConceptNet \cite{conceptnet}, CommonsenseQA \cite{commonsenseqa} aiming at generating difficult common sense questions to test models' reasoning ability. The annotators are asked to make the question as difficult as possible by looking for incorrect answers closely related to correct answers to construct the multi-choice.

COPA \cite{copa} is a task in SemEval-2012 which is a multi-choice QA dataset aiming at testing the causal reasoning ability of models. Given a premise, the models need to find the best matched cause or result. 

LogiQA \cite{logiqa} is a logical reasoning multi-choice QA dataset. The questions come from National Civil Servants Examination of China. There are mainly five kinds of reasoning types in LogiQA - categorical reasoning, sufficient conditional reasoning, necessary conditional reasoning, disjunctive reasoning and conjunctive reasoning. The question in LogiQA is extremely hard that current state-of-the-art QA model can only achieve 39\% in accuracy.

Different from existing linguistic QA dataset, OpenBookQA \cite{openbookqa} more focuses on scientific reasoning. The models should find the correct choice by reasoning between questions and the given science facts and common knowledge. 

PIQA \cite{piqa} is a special QA dataset in which the physical reasoning ability is required to solve the questions. The format of the task is select best one from two given answers. For a given scenario in question, the correct physical process needs to be chosen to achieve the expected result in question.

Given the queries from Bing as the questions and users' click, WikiQA \cite{wikiqa} selects those relevant Wikipedia pages as the context. The annotators are asked to select a correct sentence which can answer a corresponding question. Further, WikiQA also involve some questions that cannot be answered using the context.

\subsection{Domain-Specific}

CliCR \cite{clicr} is a dataset consists of gap-filling questions about medical cases, and the authors find that domain-specific knowledge is a key for success in medical QA.

BioASQ \cite{bioasq} is a extraction-based QA dataset constructed using biomedical corpus. BioASQ mainly focus on biomedical-style questions such as ``What are the physiological manifestations of disorder Y?''

The Children’s Book Test (CBT) \cite{childrensbook} is a dataset to measure the ability that language models understand children’s books. This task specifically focuses on syntactic function words prediction. 

To determine when domain-specific pre-training is worthwhile, consider current legal related NLP tasks are too easy to challenge transformer-based methods, in CaseHOLD \cite{casehold}, the authors proposed a multi-choice legal QA dataset. They use case as the context, and among some given holding statement, only one is proper. Their experiments shows that compare to normal BERT \cite{bert}, the BERT pre-trained on legal corpus can achieve better result, indicating that pre-training on domain-specific text is useful when doing difficult domain-specific tasks.

JD Production QA \cite{productqa} is a large-scale QA dataset focuses on solving the problem that how to generate a plausible answer to a question asked for a particular product in the e-commerce platform.

Mathematics Dataset \cite{methematics} is the only one QA dataset that only pay attention to mathematical reasoning, which proposed by DeepMind. With this dataset, they evaluate the sequence-to-sequence models' ability in solve mathematical problems.

PubMedQA \cite{pubmedqa} is a biomedical QA dataset which using the abstract from PubMed. For each QA-pair, the question is title or a sentence extract from a article, the article's abstract which exclude the conclusion is used as the context, and the conclusion is used as the long answer. Models should give yes/no/maybe to judge whether the conclusion can answer the question.

In QASPER \cite{qasper}, 1,585 NLP papers is used as the data source. One part of annotators are asked to give the questions based on papers' title and abstract. And the other part of annotators are asked to answer those questions base on the full text papers while also giving the corresponding support evidences.

QuAC \cite{quac} is a dialog-specific QA dataset. Based on the context provided by Wikipedia, a dialog between a teacher and a student is constructed. Different from normal QA, the QA in dialog have some differences. The question can usually be open-ended, sometimes there are no answer to a question, and the answers can be meaningful only in current dialog.

The goal of QUASAR \cite{quasart} is to test QA models' ability to first retrieve and then generate answers from retrieved documents. It consists of two sub-datasets. The first one is QUASAR-S which constructed from StackOverflow. User input queries are used to generate fill-in-the-gap questions, and users' posts and comments are used as context. The second one is QUASAR-T which consists of trivia questions and the answers come from internet.

\subsection{Multi-hop}

ComplexWebQuestions \cite{complexwebquestions} uses the simple questions from WebQuestionSP \cite{webquestionsp} build more complex questions which need to be answered by decompose the question into two simple questions. For example, the original question is ``What movies have robert pattinson starred in?'', and the complex question enhanced by SPARQL is ``What movies have robert pattinson starred in and that was produced by Erwin Stoff?''.

In ConditionalQA \cite{conditionalqa}, a correct answer is applicable only in a certain condition. Thus, the correctly answer questions, the models need to not only give the answer, but also generate the condition. The authors find that for the existing QA models, compare to generate correct answer, how to find the suitable condition is particularly challenging.

HotpotQA \cite{hotpotqa} is the first multi-hop QA dataset and the questions can only be answered by doing inference in multiple documents. Particularly, all the questions in HotpotQA are not depended on any existing knowledge and only the reasoning ability is required to generate the answer. To make the reasoning procedure more reliable, the useful sentence-level facts are also provided to help researches check whether models generate answers using the correct sources.  

While previous QA dataset only using single context, the questions in DROP \cite{drop}, however, can only be answered by get multiple fact from different paragraphs and then doing some combine operation to drive the final answer.

Compare to previous multi-hop QA only focuses on textual data, FinQA \cite{finqa}, which extracted from financial reports,  includes more kinds of information, such as table and number, and therefore some specific numerical operation is needed.

MultiRC \cite{multirc} is reading comprehension dataset where the multiple sentences should be taken into consideration to select the correct choice. It covers seven domains: science, news, travel guides, fiction stories, Wikipedia, history, 9/11 reports.

To solve the questions in WikiHop \cite{wikihop}, models need to infer multiple documents from Wikipedia to obtain the correct answers. To answer a question, models should first retrieve some support documents, and then find and combine some evident to infer the answer.

\section{Image Question Answering}

\begin{table*}[ht]
\centering
\small
\begin{tabular}{lllll}
\hline
\textbf{Dataset} & \textbf{Answer Type} & \textbf{Size} & \textbf{Domain} & \textbf{Evaluate Ability} \\ \hline
\multicolumn{5}{c}{\textbf{Image Question Answering}} \\\hline
CLEVR \cite{clevr} & Open-ended & 853K & 3D CG & Reasoning \\
RecipeQA \cite{recipeqa} & Multi-choice & 36K & Cooking Recipes & Procedural \\
CRIC \cite{cric} & Open-ended & 494K & Visual Genome & Scene Reasoning \\
DocVQA \cite{docvqa} & Open-ended & 50,000 & Document & Recognition \\
FVQA \cite{fvqa} & Open-ended & 5,826 & Open Domain & Knowledge \\ 
Visual Genome \cite{visualgenome} & Open-ended & 1,445,322 & Open Domain & Recognition \\
VCR \cite{vcr} & Multi-choice & 290K & Movie & Reasoning \\
GQA \cite{gqa} & Open-ended/Yes/No & 22M & Visual Genome & Reasoning \\
HowMany-QA \cite{howmanyqa} & Number & 106,356 & VG/VQA2.0 & Counting \\
TallyQA \cite{tallyqa} & Number & 287,907 & VG/COCO & Counting \\
TDIUC \cite{tdiuc} & Open-ended & 1.6M & VG/COCO & Multiple \\
TextVQA \cite{textvqa} & Open-ended & 45,336 & Open Domain & Text Recognition \\
VCOPA \cite{vcopa} & Multi-choice & 380 & Open Domain & Causality \\
Visual7W \cite{visual7w} & Open-ended/Multi-choice & 327,939 & VG & Reasoning \\
VizWiz \cite{vizwiz} & Open-ended & 31,000 & Photo & Recognition \\
VQA2.0 \cite{vqa2} & Open-ended & 1.11M & COCO & Recognition \\
KVQA \cite{kvqa} & Open-ended & 183,007 & Wikipedia & Knowledge \\
OK-VQA \cite{okvqa} & Open-ended & 14,000 & Open Domain & Knowledge \\
R-VQA \cite{rvqa} & Open-ended & 478,287 & VG & Reasoning \\
KB-VQA \cite{kbvqa} & Open-ended & 2,402 & COCO/ImageNet & Knowledge \\
WebQA \cite{webqa} & Open-ended & 25K & Wikipedia & Multi-hop \\
AQUA \cite{aqua} & Open-ended & 79,848 & Art & Knowledge \\
\hline
\multicolumn{5}{c}{\textbf{Video Question Answering}} \\\hline
CLEVRER \cite{clevrer} & Multi-choice & 300,000 & CG Video & Causal Reasoning \\
KnowIT VQA \cite{knowit} & Open-ended & 24,282 & TV series & Knowledge \\
MovieQA \cite{movieqa} & Open-ended & 14,944 & Movie & Reasoning \\
TVQA \cite{tvqa} & Multi-choice & 152,545 & Movie & Reasoning \\
PororoQA \cite{pororoqa} & Multi-choice & 8,913 & Cartoon & Summarization \\
Social-IQ \cite{socialiq} & Multi-choice & 7,500 & Social & Reasoning \\
\hline
\end{tabular}
\caption{Statistics of image and video QA datasets. \textit{RC} means \textit{Reading Comprehension}, \textit{MC} means \textit{Multi-choice}.}
\label{table:qadataset}
\end{table*}

\subsection{Recognition}

DocVQA \cite{docvqa} collects massive images from documents as the question context and ask questions about the textual information in the images. the models need to correctly recognize the text to generate the answer, which makes it huge different from normal object detection tasks.

HowMany-QA \cite{howmanyqa} mainly focuses on testing the counting ability of models. For each image, some counting questions are asked, such as ``How many people are wearing blue shorts?'' The images the authors used are from VQA2.0 \cite{vqa} and Visual Genome \cite{visualgenome}.

Compare to HowManyQA \cite{howmanyqa}, the counting questions in TallyQA \cite{tallyqa} are more complex and require reasoning ability about the relationship between objects and attribution. Given a simple question ``How many girraffes are there?'', the complex can be ``How many girraffes are sitting down?''. 

In TDIUC \cite{tdiuc}, there are totally 12 tasks including Counting, Scene Classification, Sentiment Understanding, etc. The main goal of TDIUC is to provide a solid benchmark to test all the existing VQA algorithms.

While previous VQA datasets merely focus on object detection, TextVQA \cite{textvqa} however, aims at the text recognition in images. Give an image, the models need to give a correct text as the answer that contained in the image to correctly answer the corresponding question.

Previous datasets only using the global information between images and QA pairs, which may ignoring some important local region information. Thus in Visual7W \cite{visual7w}, the authors first using bounding-box to extract the important objects or regions in a image, and then ask some questions about those special area.

The images in VizWiz \cite{vizwiz} are all took by blind people with a recorded spoken question. And the answers are annotated by crowdsources. This dataset is mainly used to help the research in using VQA technologies to assist blind people.

VQA2.0 \cite{vqa2} is an improved version of VQA \cite{vqa}. The questions in those two dataset are the same, asking questions about the object in the image. But the data distribution in VQA2.0 is more balanced. For each question, there are two similar images with different answers.

\subsection{Reasoning}

CLEVR \cite{clevr} aims at testing the reasoning abilities of models. Given the strong bias existing in some reasoning VQA datasets, they use computer generated 3D shapes as the image context and automatically generate some question related to attribute, counting, comparison, logical operations, etc.

Unlike other QA datasets where context is a complete paragraph, context is a procedure in RecipeQA \cite{recipeqa}. The model needs to understand the sequence of the different steps involved in making a dish in order to answer questions correctly.

While previous VQA datasets only using single object in an image, CRIC \cite{cric}, however, using the Scene Graph from Visual Genome \cite{visualgenome} to generate questions about the relationship between different objects in a same image. For example, the question can be ``What utensil can be used for moving the food that is in the bowl?'' and models are asked to find the correct object as the answer that is spoon. CRIC challenges the relational reasoning ability of visual models.

Visual Genome \cite{visualgenome} is the largest VQA dataset. The questions in Visual Genome always starts with six Ws - what, where, when, who, why, and how, which allow them to test wide range of model abilities such as detection, categorization, commonsense reasoning. Using the annotated scene graphs, human workers are asked to give questions based on region descriptions.

VCR \cite{vcr} combines object recognition and relation reasoning through asking questions between different objects. For example, given a image contains some people which are already highlighted and annotated by bounding boxes, the question can be ``Why the \texttt{[person1]} talk to \texttt{[person2]}?'' To correctly answer this question, models should first select the most rational answer from four answers and then should find the most reasonable explanation also from four candidates to explain why the first choice is correct.

GQA \cite{gqa} leverage the scene graph provided by Visual Genome \cite{visualgenome} to generate questions about the objects in image. The question involves object recognition, classification of relations between objects and so on, such as  ``Is the bowl to the right of the green apple?'' They use different templates to automatically generate questions and thus the total number of questions achieves 22M.

Inspired by COPA \cite{copa}, VCOPA \cite{vcopa} researches the causality in VQA scenario. Given an image as the premise, the VQA models need to choose the most possible result that may caused by this premise from two images.

R-VQA \cite{rvqa} mainly aims at the relationship reasoning between different entities. They choose three types of relations: \texttt{(there, is, object)}, \texttt{(subject, is, attribute)}, \texttt{(subject, relation, object)}. For each QA pair, they also provide the corresponding relation to help answer the question.

\subsection{Commonsense and Knowledge}

Different from previous datasets that questions can be answered merely rely on the analysis of image itself, the questions in FVQA \cite{fvqa} can only be solved considering some external commonsense and knowledge. This makes the dataset can be easily answered by human but much difficult to models. For instance, given a image contains a red fire hydrant, the question can be ``What can the red object on the ground be used for ?'' and the correct answer is ``Firefighting''.

KVQA \cite{kvqa} focuses on the world knowledge about the famous people. They collect the images in Wikipedia and asking attribute questions, such as  ``who is the person in the images?'' or ``How old is the person?''

For each sample in OK-VQA \cite{okvqa}, in addition to a question and an answer, they also provide some outside knowledge, in which the correct answer contained. Those knowledge a closely related to the QA pair, and model should analysis both the question and outside knowledge to extract the final answer.

Instead of merely asking questions about visual, questions in KB-VQA \cite{kbvqa} may also related to commonsense and knowledge. For example, if the visual question is ``How many giraffes are there in the image'', then the commonsense question can be ``Is this image related to zoology?'', and the corresponding knowledge question is ``What are the common properties between the animal in this image and zebra?''.

AQUA \cite{aqua} is a VQA dataset specifically about Art. The questions are generate in two different ways. The first is directly generated based on the painting itself, which can be answered only using the painting. And the other way is to generate from the painting's comments which may involve some external knowledge and cannot be answer merely rely on the painting.

\subsection{Multi-hop}

WebQA \cite{webqa} is currently the only one multi-hop VQA dataset. To ask the questions, the models need to reasoning on some images and text snippets. Some images and text are related to the question while some not. Models should first find the related images and text then combine them together to generate the correct answer.

\section{Video Question Answering}

CLEVRER \cite{clevrer} focuses on the temporal causal reasoning abilities of models. The authors use CG technology to simulate the movement and collision between different simple geometric objects to design questions and answers. Asking questions such as ``Which of the following is responsible for the gray cylinder’s colliding with the cube?'' and the models need to figure out the correct reason for this.

KnowIT VQA \cite{knowit} use the clips from \textit{The Big Bang Theory} to generate questions. There totally four kinds of questions: visual, textual, temporal and knowledge. Models should use both the video and the subtitles, as well as some external knowledge about the TV series, to answer the questions.

There are there types of source information in MovieQA \cite{movieqa}: plot, video and subtitle. Combine those different information, models should answer the annotated questions such as ``Who kills Neo in the Matrix?''.

In TVQA \cite{tvqa}, some questions can be answer by using video or subtitle independently while some questions can only be answered by combining both of them. Furthermore, the models should also learn to precisely locate where the useful clips are.

PororoQA \cite{pororoqa} is a VQA dataset specially focusing on cartoon for children and thus the word in it are quite simple. There are totally ten types of questions: Action, Person, Abstract, Detail, Method, Reason, Location, Statement, Causality, Yes/No, Time. 

Social-IQ \cite{socialiq} aims to help the AI researches in social situation. Given some video scenes where people socialize, the annotators are asked to ask some question related to those people. For example, ``How is the discussion between the woman and the man in the white shirt?'' 

\section{Challenges and Chances}

A comparison of those three different forms of QA shows that, given how early they were introduced, textual QA is the most studied among the three QA forms, which has the most number of datasets, and covers the most types of QA abilities. 
For VQA, because it is just proposed in the new era of deep learning, there is still a lack of research and datasets, especially for video QA. 
This is also closely related to the difficulty of corpus collection, annotation difficulty and training resource demand.

\subsection{Textual QA}

Because textual information are most available in real-world or online websites, Textual QA is both the earliest and the most studied direction. 
There are two main sources of textual QA today. One is a variety of actual exam questions, and one is human design questions from the annotator. 
Questions from exams are more difficult and of higher quality, but are also limited in quantity to meet the needs of each particular type of QA. 
On the contrary, Questions designed by human annotators are more flexible and thus can meet different demands, but the annotate process are expensive and usually contains some errors which cannot measure up to the precision of the exam questions. 

Although various datasets are now available, there are still many unsolved problems. 
The first is the evaluation matrix of QA model performance. QA is currently measured primarily by the overlap degree to which a predicted answer meets a target answer, and the higher the overlap, the more accurate it is considered. But language is complex, and there can be more than one correct answer to the same question, and word coincidence doesn't take actual semantics into consideration, which may lead to some misjudgments. 
The second is the interpretability of QA models. The current QA models just give the answer directly, we cannot know how the answer is extracted or generated, the whole process is unexplainable.
The third is the gap between QA datasets and the real-world scenario. Out of domain problem is a critical disadvantage for all deep learning-related research, and QA is no exception. Although some models currently achieve more than 90\% accuracy on some particular datasets, there is always a huge performance degradation when they are applied to other fields or practical applications. How to narrow this gap is also a problem worth studying.
And the last is the limitation of question types. In the current QA datasets, questions mainly start with \textit{what}, \textit{when}, \textit{where}, etc., while there are few open-end questions like \textit{why} and \textit{how}. The answer to the former is always short and easily solved by the model, while the answer to the latter may be much longer than the question itself. This is a big challenge for capacity-limited deep learning models. We can't use infinite huge models because of the cost of machines, but knowledge itself is infinite.

\subsection{Visual QA}

Visual QA, including image QA and video QA, is in general an extension of textual QA. The major difference is in VQA, the context is image or video rather than text. In addition to having some of the same issues as textual QA, VQA has some unique problems.
Firstly, images and videos are more difficult to access and annotate than text, making the number of VQA datasets much smaller than those of textual QA, and usually with lower quality.
Secondly, in QA tasks, there is a higher requirement for understanding images. Traditional computer vision tasks are mainly used for object classification or object recognition. But to answer the question, the model needs to recognize not only the object, but also the relationship between the objects and even the meaning of the whole image.
And lastly, compared with textual QA, there are still many problems to be studied in VQA due to its late start. Luckily, many studies in textual QA can be transferred to VQA and provide some references for VQA.

\section{Conclusion}

In this work, we investigate and present a comprehensive survey on QA. Specifically we focuses on the QA datasets of three different QA tasks: textual QA, image QA and video QA. For each dataset, we introduce its data statistics and usage. Moreover, through self-comparison and cross-comparison of these three tasks, we analyze some of the current challenges and opportunities QA research facing, to provide some reference and insight for the future QA researches.

\bibliography{anthology,custom}

\begin{thebibliography}{75}
\expandafter\ifx\csname natexlab\endcsname\relax\def\natexlab#1{#1}\fi

\bibitem[{Acharya et~al.(2019)Acharya, Kafle, and Kanan}]{tallyqa}
Manoj Acharya, Kushal Kafle, and Christopher Kanan. 2019.
\newblock Tallyqa: Answering complex counting questions.
\newblock In \emph{Proceedings of the AAAI Conference on Artificial
  Intelligence}, volume~33, pages 8076--8084.

\bibitem[{Antol et~al.(2015)Antol, Agrawal, Lu, Mitchell, Batra, Zitnick, and
  Parikh}]{vqa}
Stanislaw Antol, Aishwarya Agrawal, Jiasen Lu, Margaret Mitchell, Dhruv Batra,
  C~Lawrence Zitnick, and Devi Parikh. 2015.
\newblock Vqa: Visual question answering.
\newblock In \emph{Proceedings of the IEEE international conference on computer
  vision}, pages 2425--2433.

\bibitem[{Bisk et~al.(2020)Bisk, Zellers, Gao, Choi et~al.}]{piqa}
Yonatan Bisk, Rowan Zellers, Jianfeng Gao, Yejin Choi, et~al. 2020.
\newblock Piqa: Reasoning about physical commonsense in natural language.
\newblock In \emph{Proceedings of the AAAI Conference on Artificial
  Intelligence}, volume~34, pages 7432--7439.

\bibitem[{Bordes et~al.(2015)Bordes, Usunier, Chopra, and
  Weston}]{simplequestions}
Antoine Bordes, Nicolas Usunier, Sumit Chopra, and Jason Weston. 2015.
\newblock Large-scale simple question answering with memory networks.
\newblock \emph{arXiv preprint arXiv:1506.02075}.

\bibitem[{Chang et~al.(2021)Chang, Narang, Suzuki, Cao, Gao, and Bisk}]{webqa}
Yingshan Chang, Mridu Narang, Hisami Suzuki, Guihong Cao, Jianfeng Gao, and
  Yonatan Bisk. 2021.
\newblock Webqa: Multihop and multimodal qa.
\newblock \emph{arXiv preprint arXiv:2109.00590}.

\bibitem[{Chen et~al.(2019)Chen, D’Arcy, Liu, Fernandez, and Downey}]{codah}
Michael Chen, Mike D’Arcy, Alisa Liu, Jared Fernandez, and Doug Downey. 2019.
\newblock Codah: An adversarially-authored question answering dataset for
  common sense.
\newblock In \emph{Proceedings of the 3rd Workshop on Evaluating Vector Space
  Representations for NLP}, pages 63--69.

\bibitem[{Chen et~al.(2021)Chen, Chen, Smiley, Shah, Borova, Langdon, Moussa,
  Beane, Huang, Routledge et~al.}]{finqa}
Zhiyu Chen, Wenhu Chen, Charese Smiley, Sameena Shah, Iana Borova, Dylan
  Langdon, Reema Moussa, Matt Beane, Ting-Hao Huang, Bryan Routledge, et~al.
  2021.
\newblock Finqa: A dataset of numerical reasoning over financial data.
\newblock \emph{arXiv preprint arXiv:2109.00122}.

\bibitem[{Choi et~al.(2018)Choi, He, Iyyer, Yatskar, Yih, Choi, Liang, and
  Zettlemoyer}]{quac}
Eunsol Choi, He~He, Mohit Iyyer, Mark Yatskar, Wen-tau Yih, Yejin Choi, Percy
  Liang, and Luke Zettlemoyer. 2018.
\newblock Quac: Question answering in context.
\newblock \emph{arXiv preprint arXiv:1808.07036}.

\bibitem[{Clark et~al.(2019)Clark, Lee, Chang, Kwiatkowski, Collins, and
  Toutanova}]{boolq}
Christopher Clark, Kenton Lee, Ming-Wei Chang, Tom Kwiatkowski, Michael
  Collins, and Kristina Toutanova. 2019.
\newblock Boolq: Exploring the surprising difficulty of natural yes/no
  questions.
\newblock \emph{arXiv preprint arXiv:1905.10044}.

\bibitem[{Clark et~al.(2018)Clark, Cowhey, Etzioni, Khot, Sabharwal, Schoenick,
  and Tafjord}]{arc}
Peter Clark, Isaac Cowhey, Oren Etzioni, Tushar Khot, Ashish Sabharwal, Carissa
  Schoenick, and Oyvind Tafjord. 2018.
\newblock Think you have solved question answering? try arc, the ai2 reasoning
  challenge.
\newblock \emph{arXiv preprint arXiv:1803.05457}.

\bibitem[{Dasigi et~al.(2021)Dasigi, Lo, Beltagy, Cohan, Smith, and
  Gardner}]{qasper}
Pradeep Dasigi, Kyle Lo, Iz~Beltagy, Arman Cohan, Noah~A Smith, and Matt
  Gardner. 2021.
\newblock A dataset of information-seeking questions and answers anchored in
  research papers.
\newblock \emph{arXiv preprint arXiv:2105.03011}.

\bibitem[{Devlin et~al.(2018)Devlin, Chang, Lee, and Toutanova}]{bert}
Jacob Devlin, Ming-Wei Chang, Kenton Lee, and Kristina Toutanova. 2018.
\newblock Bert: Pre-training of deep bidirectional transformers for language
  understanding.
\newblock \emph{arXiv preprint arXiv:1810.04805}.

\bibitem[{Dhingra et~al.(2017)Dhingra, Mazaitis, and Cohen}]{quasart}
Bhuwan Dhingra, Kathryn Mazaitis, and William~W Cohen. 2017.
\newblock Quasar: Datasets for question answering by search and reading.
\newblock \emph{arXiv preprint arXiv:1707.03904}.

\bibitem[{Dua et~al.(2019)Dua, Wang, Dasigi, Stanovsky, Singh, and
  Gardner}]{drop}
Dheeru Dua, Yizhong Wang, Pradeep Dasigi, Gabriel Stanovsky, Sameer Singh, and
  Matt Gardner. 2019.
\newblock Drop: A reading comprehension benchmark requiring discrete reasoning
  over paragraphs.
\newblock \emph{arXiv preprint arXiv:1903.00161}.

\bibitem[{Gao et~al.(2019{\natexlab{a}})Gao, Wang, Shan, and Chen}]{cric}
Difei Gao, Ruiping Wang, Shiguang Shan, and Xilin Chen. 2019{\natexlab{a}}.
\newblock From two graphs to n questions: A vqa dataset for compositional
  reasoning on vision and commonsense.
\newblock \emph{arXiv preprint arXiv:1908.02962}.

\bibitem[{Gao et~al.(2019{\natexlab{b}})Gao, Ren, Zhao, Zhao, Yin, and
  Yan}]{productqa}
Shen Gao, Zhaochun Ren, Yihong Zhao, Dongyan Zhao, Dawei Yin, and Rui Yan.
  2019{\natexlab{b}}.
\newblock Product-aware answer generation in e-commerce question-answering.
\newblock In \emph{Proceedings of the Twelfth ACM International Conference on
  Web Search and Data Mining}, pages 429--437.

\bibitem[{Garcia et~al.(2020{\natexlab{a}})Garcia, Otani, Chu, and
  Nakashima}]{knowit}
Noa Garcia, Mayu Otani, Chenhui Chu, and Yuta Nakashima. 2020{\natexlab{a}}.
\newblock Knowit vqa: Answering knowledge-based questions about videos.
\newblock In \emph{Proceedings of the AAAI Conference on Artificial
  Intelligence}, volume~34, pages 10826--10834.

\bibitem[{Garcia et~al.(2020{\natexlab{b}})Garcia, Ye, Liu, Hu, Otani, Chu,
  Nakashima, and Mitamura}]{aqua}
Noa Garcia, Chentao Ye, Zihua Liu, Qingtao Hu, Mayu Otani, Chenhui Chu, Yuta
  Nakashima, and Teruko Mitamura. 2020{\natexlab{b}}.
\newblock A dataset and baselines for visual question answering on art.
\newblock In \emph{European Conference on Computer Vision}, pages 92--108.
  Springer.

\bibitem[{Gordon et~al.(2012)Gordon, Kozareva, and Roemmele}]{copa}
Andrew Gordon, Zornitsa Kozareva, and Melissa Roemmele. 2012.
\newblock Semeval-2012 task 7: Choice of plausible alternatives: An evaluation
  of commonsense causal reasoning.
\newblock In \emph{* SEM 2012: The First Joint Conference on Lexical and
  Computational Semantics--Volume 1: Proceedings of the main conference and the
  shared task, and Volume 2: Proceedings of the Sixth International Workshop on
  Semantic Evaluation (SemEval 2012)}, pages 394--398.

\bibitem[{Goyal et~al.(2017)Goyal, Khot, Summers-Stay, Batra, and
  Parikh}]{vqa2}
Yash Goyal, Tejas Khot, Douglas Summers-Stay, Dhruv Batra, and Devi Parikh.
  2017.
\newblock Making the v in vqa matter: Elevating the role of image understanding
  in visual question answering.
\newblock In \emph{Proceedings of the IEEE Conference on Computer Vision and
  Pattern Recognition}, pages 6904--6913.

\bibitem[{Gurari et~al.(2018)Gurari, Li, Stangl, Guo, Lin, Grauman, Luo, and
  Bigham}]{vizwiz}
Danna Gurari, Qing Li, Abigale~J Stangl, Anhong Guo, Chi Lin, Kristen Grauman,
  Jiebo Luo, and Jeffrey~P Bigham. 2018.
\newblock Vizwiz grand challenge: Answering visual questions from blind people.
\newblock In \emph{Proceedings of the IEEE Conference on Computer Vision and
  Pattern Recognition}, pages 3608--3617.

\bibitem[{Hill et~al.(2015)Hill, Bordes, Chopra, and Weston}]{childrensbook}
Felix Hill, Antoine Bordes, Sumit Chopra, and Jason Weston. 2015.
\newblock The goldilocks principle: Reading children's books with explicit
  memory representations.
\newblock \emph{arXiv preprint arXiv:1511.02301}.

\bibitem[{Hirschman and Gaizauskas(2001)}]{qa}
Lynette Hirschman and Robert Gaizauskas. 2001.
\newblock Natural language question answering: the view from here.
\newblock \emph{natural language engineering}, 7(4):275--300.

\bibitem[{Hudson and Manning(2019)}]{gqa}
Drew~A Hudson and Christopher~D Manning. 2019.
\newblock Gqa: A new dataset for real-world visual reasoning and compositional
  question answering.
\newblock In \emph{Proceedings of the IEEE/CVF conference on computer vision
  and pattern recognition}, pages 6700--6709.

\bibitem[{Jin et~al.(2019)Jin, Dhingra, Liu, Cohen, and Lu}]{pubmedqa}
Qiao Jin, Bhuwan Dhingra, Zhengping Liu, William~W Cohen, and Xinghua Lu. 2019.
\newblock Pubmedqa: A dataset for biomedical research question answering.
\newblock \emph{arXiv preprint arXiv:1909.06146}.

\bibitem[{Johnson et~al.(2017)Johnson, Hariharan, Van Der~Maaten, Fei-Fei,
  Lawrence~Zitnick, and Girshick}]{clevr}
Justin Johnson, Bharath Hariharan, Laurens Van Der~Maaten, Li~Fei-Fei,
  C~Lawrence~Zitnick, and Ross Girshick. 2017.
\newblock Clevr: A diagnostic dataset for compositional language and elementary
  visual reasoning.
\newblock In \emph{Proceedings of the IEEE conference on computer vision and
  pattern recognition}, pages 2901--2910.

\bibitem[{Joshi et~al.(2017)Joshi, Choi, Weld, and Zettlemoyer}]{triviaqa}
Mandar Joshi, Eunsol Choi, Daniel~S Weld, and Luke Zettlemoyer. 2017.
\newblock Triviaqa: A large scale distantly supervised challenge dataset for
  reading comprehension.
\newblock \emph{arXiv preprint arXiv:1705.03551}.

\bibitem[{Kafle and Kanan(2017)}]{tdiuc}
Kushal Kafle and Christopher Kanan. 2017.
\newblock An analysis of visual question answering algorithms.
\newblock In \emph{Proceedings of the IEEE International Conference on Computer
  Vision}, pages 1965--1973.

\bibitem[{Khashabi et~al.(2018)Khashabi, Chaturvedi, Roth, Upadhyay, and
  Roth}]{multirc}
Daniel Khashabi, Snigdha Chaturvedi, Michael Roth, Shyam Upadhyay, and Dan
  Roth. 2018.
\newblock Looking beyond the surface: A challenge set for reading comprehension
  over multiple sentences.
\newblock In \emph{Proceedings of the 2018 Conference of the North American
  Chapter of the Association for Computational Linguistics: Human Language
  Technologies, Volume 1 (Long Papers)}, pages 252--262.

\bibitem[{Kim et~al.(2017)Kim, Heo, Choi, and Zhang}]{pororoqa}
Kyung-Min Kim, Min-Oh Heo, Seong-Ho Choi, and Byoung-Tak Zhang. 2017.
\newblock Deepstory: Video story qa by deep embedded memory networks.
\newblock \emph{arXiv preprint arXiv:1707.00836}.

\bibitem[{Ko{\v{c}}isk{\`y} et~al.(2018)Ko{\v{c}}isk{\`y}, Schwarz, Blunsom,
  Dyer, Hermann, Melis, and Grefenstette}]{narrativeqa}
Tom{\'a}{\v{s}} Ko{\v{c}}isk{\`y}, Jonathan Schwarz, Phil Blunsom, Chris Dyer,
  Karl~Moritz Hermann, G{\'a}bor Melis, and Edward Grefenstette. 2018.
\newblock The narrativeqa reading comprehension challenge.
\newblock \emph{Transactions of the Association for Computational Linguistics},
  6:317--328.

\bibitem[{Kong et~al.(2020)Kong, Gangal, and Hovy}]{scde}
Xiang Kong, Varun Gangal, and Eduard Hovy. 2020.
\newblock Scde: sentence cloze dataset with high quality distractors from
  examinations.
\newblock \emph{arXiv preprint arXiv:2004.12934}.

\bibitem[{Krishna et~al.(2017)Krishna, Zhu, Groth, Johnson, Hata, Kravitz,
  Chen, Kalantidis, Li, Shamma et~al.}]{visualgenome}
Ranjay Krishna, Yuke Zhu, Oliver Groth, Justin Johnson, Kenji Hata, Joshua
  Kravitz, Stephanie Chen, Yannis Kalantidis, Li-Jia Li, David~A Shamma, et~al.
  2017.
\newblock Visual genome: Connecting language and vision using crowdsourced
  dense image annotations.
\newblock \emph{International journal of computer vision}, 123(1):32--73.

\bibitem[{Kwiatkowski et~al.(2019)Kwiatkowski, Palomaki, Redfield, Collins,
  Parikh, Alberti, Epstein, Polosukhin, Devlin, Lee et~al.}]{naturalquestions}
Tom Kwiatkowski, Jennimaria Palomaki, Olivia Redfield, Michael Collins, Ankur
  Parikh, Chris Alberti, Danielle Epstein, Illia Polosukhin, Jacob Devlin,
  Kenton Lee, et~al. 2019.
\newblock Natural questions: a benchmark for question answering research.
\newblock \emph{Transactions of the Association for Computational Linguistics},
  7:453--466.

\bibitem[{Lai et~al.(2017)Lai, Xie, Liu, Yang, and Hovy}]{race}
Guokun Lai, Qizhe Xie, Hanxiao Liu, Yiming Yang, and Eduard Hovy. 2017.
\newblock Race: Large-scale reading comprehension dataset from examinations.
\newblock \emph{arXiv preprint arXiv:1704.04683}.

\bibitem[{Lei et~al.(2018)Lei, Yu, Bansal, and Berg}]{tvqa}
Jie Lei, Licheng Yu, Mohit Bansal, and Tamara~L Berg. 2018.
\newblock Tvqa: Localized, compositional video question answering.
\newblock \emph{arXiv preprint arXiv:1809.01696}.

\bibitem[{Liu and Singh(2004)}]{conceptnet}
Hugo Liu and Push Singh. 2004.
\newblock Conceptnet—a practical commonsense reasoning tool-kit.
\newblock \emph{BT technology journal}, 22(4):211--226.

\bibitem[{Liu et~al.(2020)Liu, Cui, Liu, Huang, Wang, and Zhang}]{logiqa}
Jian Liu, Leyang Cui, Hanmeng Liu, Dandan Huang, Yile Wang, and Yue Zhang.
  2020.
\newblock Logiqa: A challenge dataset for machine reading comprehension with
  logical reasoning.
\newblock \emph{arXiv preprint arXiv:2007.08124}.

\bibitem[{Lu et~al.(2018)Lu, Ji, Zhang, Duan, Zhou, and Wang}]{rvqa}
Pan Lu, Lei Ji, Wei Zhang, Nan Duan, Ming Zhou, and Jianyong Wang. 2018.
\newblock R-vqa: learning visual relation facts with semantic attention for
  visual question answering.
\newblock In \emph{Proceedings of the 24th ACM SIGKDD International Conference
  on Knowledge Discovery \& Data Mining}, pages 1880--1889.

\bibitem[{Marino et~al.(2019)Marino, Rastegari, Farhadi, and Mottaghi}]{okvqa}
Kenneth Marino, Mohammad Rastegari, Ali Farhadi, and Roozbeh Mottaghi. 2019.
\newblock Ok-vqa: A visual question answering benchmark requiring external
  knowledge.
\newblock In \emph{Proceedings of the IEEE/CVF Conference on Computer Vision
  and Pattern Recognition}, pages 3195--3204.

\bibitem[{Mathew et~al.(2021)Mathew, Karatzas, and Jawahar}]{docvqa}
Minesh Mathew, Dimosthenis Karatzas, and CV~Jawahar. 2021.
\newblock Docvqa: A dataset for vqa on document images.
\newblock In \emph{Proceedings of the IEEE/CVF Winter Conference on
  Applications of Computer Vision}, pages 2200--2209.

\bibitem[{Mihaylov et~al.(2018)Mihaylov, Clark, Khot, and
  Sabharwal}]{openbookqa}
Todor Mihaylov, Peter Clark, Tushar Khot, and Ashish Sabharwal. 2018.
\newblock Can a suit of armor conduct electricity? a new dataset for open book
  question answering.
\newblock \emph{arXiv preprint arXiv:1809.02789}.

\bibitem[{Nguyen et~al.(2016)Nguyen, Rosenberg, Song, Gao, Tiwary, Majumder,
  and Deng}]{msmarco}
Tri Nguyen, Mir Rosenberg, Xia Song, Jianfeng Gao, Saurabh Tiwary, Rangan
  Majumder, and Li~Deng. 2016.
\newblock Ms marco: A human generated machine reading comprehension dataset.
\newblock In \emph{CoCo@ NIPS}.

\bibitem[{Rajpurkar et~al.(2018)Rajpurkar, Jia, and Liang}]{squad2}
Pranav Rajpurkar, Robin Jia, and Percy Liang. 2018.
\newblock Know what you don't know: Unanswerable questions for squad.
\newblock \emph{arXiv preprint arXiv:1806.03822}.

\bibitem[{Rajpurkar et~al.(2016)Rajpurkar, Zhang, Lopyrev, and Liang}]{squad1}
Pranav Rajpurkar, Jian Zhang, Konstantin Lopyrev, and Percy Liang. 2016.
\newblock Squad: 100,000+ questions for machine comprehension of text.
\newblock \emph{arXiv preprint arXiv:1606.05250}.

\bibitem[{Reddy et~al.(2019)Reddy, Chen, and Manning}]{coqa}
Siva Reddy, Danqi Chen, and Christopher~D Manning. 2019.
\newblock Coqa: A conversational question answering challenge.
\newblock \emph{Transactions of the Association for Computational Linguistics},
  7:249--266.

\bibitem[{Richardson et~al.(2013)Richardson, Burges, and Renshaw}]{mctest}
Matthew Richardson, Christopher~JC Burges, and Erin Renshaw. 2013.
\newblock Mctest: A challenge dataset for the open-domain machine comprehension
  of text.
\newblock In \emph{Proceedings of the 2013 conference on empirical methods in
  natural language processing}, pages 193--203.

\bibitem[{Saxton et~al.(2019)Saxton, Grefenstette, Hill, and
  Kohli}]{methematics}
David Saxton, Edward Grefenstette, Felix Hill, and Pushmeet Kohli. 2019.
\newblock Analysing mathematical reasoning abilities of neural models.
\newblock \emph{arXiv preprint arXiv:1904.01557}.

\bibitem[{See et~al.(2017)See, Liu, and Manning}]{cnndaily}
Abigail See, Peter~J Liu, and Christopher~D Manning. 2017.
\newblock Get to the point: Summarization with pointer-generator networks.
\newblock \emph{arXiv preprint arXiv:1704.04368}.

\bibitem[{Shah et~al.(2019)Shah, Mishra, Yadati, and Talukdar}]{kvqa}
Sanket Shah, Anand Mishra, Naganand Yadati, and Partha~Pratim Talukdar. 2019.
\newblock Kvqa: Knowledge-aware visual question answering.
\newblock In \emph{Proceedings of the AAAI Conference on Artificial
  Intelligence}, volume~33, pages 8876--8884.

\bibitem[{Singh et~al.(2019)Singh, Natarajan, Shah, Jiang, Chen, Batra, Parikh,
  and Rohrbach}]{textvqa}
Amanpreet Singh, Vivek Natarajan, Meet Shah, Yu~Jiang, Xinlei Chen, Dhruv
  Batra, Devi Parikh, and Marcus Rohrbach. 2019.
\newblock Towards vqa models that can read.
\newblock In \emph{Proceedings of the IEEE/CVF Conference on Computer Vision
  and Pattern Recognition}, pages 8317--8326.

\bibitem[{Sun et~al.(2021)Sun, Cohen, and Salakhutdinov}]{conditionalqa}
Haitian Sun, William~W Cohen, and Ruslan Salakhutdinov. 2021.
\newblock Conditionalqa: A complex reading comprehension dataset with
  conditional answers.
\newblock \emph{arXiv preprint arXiv:2110.06884}.

\bibitem[{{\v{S}}uster and Daelemans(2018)}]{clicr}
Simon {\v{S}}uster and Walter Daelemans. 2018.
\newblock Clicr: A dataset of clinical case reports for machine reading
  comprehension.
\newblock \emph{arXiv preprint arXiv:1803.09720}.

\bibitem[{Talmor and Berant(2018)}]{complexwebquestions}
Alon Talmor and Jonathan Berant. 2018.
\newblock The web as a knowledge-base for answering complex questions.
\newblock \emph{arXiv preprint arXiv:1803.06643}.

\bibitem[{Talmor et~al.(2018)Talmor, Herzig, Lourie, and
  Berant}]{commonsenseqa}
Alon Talmor, Jonathan Herzig, Nicholas Lourie, and Jonathan Berant. 2018.
\newblock Commonsenseqa: A question answering challenge targeting commonsense
  knowledge.
\newblock \emph{arXiv preprint arXiv:1811.00937}.

\bibitem[{Tapaswi et~al.(2016)Tapaswi, Zhu, Stiefelhagen, Torralba, Urtasun,
  and Fidler}]{movieqa}
Makarand Tapaswi, Yukun Zhu, Rainer Stiefelhagen, Antonio Torralba, Raquel
  Urtasun, and Sanja Fidler. 2016.
\newblock Movieqa: Understanding stories in movies through question-answering.
\newblock In \emph{Proceedings of the IEEE conference on computer vision and
  pattern recognition}, pages 4631--4640.

\bibitem[{Trischler et~al.(2016)Trischler, Wang, Yuan, Harris, Sordoni,
  Bachman, and Suleman}]{newsqa}
Adam Trischler, Tong Wang, Xingdi Yuan, Justin Harris, Alessandro Sordoni,
  Philip Bachman, and Kaheer Suleman. 2016.
\newblock Newsqa: A machine comprehension dataset.
\newblock \emph{arXiv preprint arXiv:1611.09830}.

\bibitem[{Trott et~al.(2017)Trott, Xiong, and Socher}]{howmanyqa}
Alexander Trott, Caiming Xiong, and Richard Socher. 2017.
\newblock Interpretable counting for visual question answering.
\newblock \emph{arXiv preprint arXiv:1712.08697}.

\bibitem[{Tsatsaronis et~al.(2015)Tsatsaronis, Balikas, Malakasiotis, Partalas,
  Zschunke, Alvers, Weissenborn, Krithara, Petridis, Polychronopoulos
  et~al.}]{bioasq}
George Tsatsaronis, Georgios Balikas, Prodromos Malakasiotis, Ioannis Partalas,
  Matthias Zschunke, Michael~R Alvers, Dirk Weissenborn, Anastasia Krithara,
  Sergios Petridis, Dimitris Polychronopoulos, et~al. 2015.
\newblock An overview of the bioasq large-scale biomedical semantic indexing
  and question answering competition.
\newblock \emph{BMC bioinformatics}, 16(1):1--28.

\bibitem[{Wang et~al.(2017)Wang, Wu, Shen, Dick, and Van Den~Hengel}]{fvqa}
Peng Wang, Qi~Wu, Chunhua Shen, Anthony Dick, and Anton Van Den~Hengel. 2017.
\newblock Fvqa: Fact-based visual question answering.
\newblock \emph{IEEE transactions on pattern analysis and machine
  intelligence}, 40(10):2413--2427.

\bibitem[{Wang et~al.(2015)Wang, Wu, Shen, Hengel, and Dick}]{kbvqa}
Peng Wang, Qi~Wu, Chunhua Shen, Anton van~den Hengel, and Anthony Dick. 2015.
\newblock Explicit knowledge-based reasoning for visual question answering.
\newblock \emph{arXiv preprint arXiv:1511.02570}.

\bibitem[{Welbl et~al.(2018)Welbl, Stenetorp, and Riedel}]{wikihop}
Johannes Welbl, Pontus Stenetorp, and Sebastian Riedel. 2018.
\newblock Constructing datasets for multi-hop reading comprehension across
  documents.
\newblock \emph{Transactions of the Association for Computational Linguistics},
  6:287--302.

\bibitem[{Weston et~al.(2015)Weston, Bordes, Chopra, Rush, van Merri{\"e}nboer,
  Joulin, and Mikolov}]{babi}
Jason Weston, Antoine Bordes, Sumit Chopra, Alexander~M Rush, Bart van
  Merri{\"e}nboer, Armand Joulin, and Tomas Mikolov. 2015.
\newblock Towards ai-complete question answering: A set of prerequisite toy
  tasks.
\newblock \emph{arXiv preprint arXiv:1502.05698}.

\bibitem[{Xiong et~al.(2019)Xiong, Wu, Wang, Kulkarni, Yu, Chang, Guo, and
  Wang}]{tweetqa}
Wenhan Xiong, Jiawei Wu, Hong Wang, Vivek Kulkarni, Mo~Yu, Shiyu Chang,
  Xiaoxiao Guo, and William~Yang Wang. 2019.
\newblock Tweetqa: A social media focused question answering dataset.
\newblock \emph{arXiv preprint arXiv:1907.06292}.

\bibitem[{Yagcioglu et~al.(2018)Yagcioglu, Erdem, Erdem, and
  Ikizler-Cinbis}]{recipeqa}
Semih Yagcioglu, Aykut Erdem, Erkut Erdem, and Nazli Ikizler-Cinbis. 2018.
\newblock Recipeqa: A challenge dataset for multimodal comprehension of cooking
  recipes.
\newblock \emph{arXiv preprint arXiv:1809.00812}.

\bibitem[{Yang et~al.(2015)Yang, Yih, and Meek}]{wikiqa}
Yi~Yang, Wen-tau Yih, and Christopher Meek. 2015.
\newblock Wikiqa: A challenge dataset for open-domain question answering.
\newblock In \emph{Proceedings of the 2015 conference on empirical methods in
  natural language processing}, pages 2013--2018.

\bibitem[{Yang et~al.(2018)Yang, Qi, Zhang, Bengio, Cohen, Salakhutdinov, and
  Manning}]{hotpotqa}
Zhilin Yang, Peng Qi, Saizheng Zhang, Yoshua Bengio, William~W Cohen, Ruslan
  Salakhutdinov, and Christopher~D Manning. 2018.
\newblock Hotpotqa: A dataset for diverse, explainable multi-hop question
  answering.
\newblock \emph{arXiv preprint arXiv:1809.09600}.

\bibitem[{Yeo et~al.(2018)Yeo, Lee, Wang, Choi, Cho, Amplayo, and
  Hwang}]{vcopa}
Jinyoung Yeo, Gyeongbok Lee, Gengyu Wang, Seungtaek Choi, Hyunsouk Cho,
  Reinald~Kim Amplayo, and Seung-won Hwang. 2018.
\newblock Visual choice of plausible alternatives: An evaluation of image-based
  commonsense causal reasoning.
\newblock In \emph{Proceedings of the Eleventh International Conference on
  Language Resources and Evaluation (LREC 2018)}.

\bibitem[{Yi et~al.(2019)Yi, Gan, Li, Kohli, Wu, Torralba, and
  Tenenbaum}]{clevrer}
Kexin Yi, Chuang Gan, Yunzhu Li, Pushmeet Kohli, Jiajun Wu, Antonio Torralba,
  and Joshua~B Tenenbaum. 2019.
\newblock Clevrer: Collision events for video representation and reasoning.
\newblock \emph{arXiv preprint arXiv:1910.01442}.

\bibitem[{Yih et~al.(2016)Yih, Richardson, Meek, Chang, and
  Suh}]{webquestionsp}
Wen-tau Yih, Matthew Richardson, Christopher Meek, Ming-Wei Chang, and Jina
  Suh. 2016.
\newblock The value of semantic parse labeling for knowledge base question
  answering.
\newblock In \emph{Proceedings of the 54th Annual Meeting of the Association
  for Computational Linguistics (Volume 2: Short Papers)}, pages 201--206.

\bibitem[{Yu et~al.(2020)Yu, Jiang, Dong, and Feng}]{reclor}
Weihao Yu, Zihang Jiang, Yanfei Dong, and Jiashi Feng. 2020.
\newblock Reclor: A reading comprehension dataset requiring logical reasoning.
\newblock \emph{arXiv preprint arXiv:2002.04326}.

\bibitem[{Zadeh et~al.(2019)Zadeh, Chan, Liang, Tong, and Morency}]{socialiq}
Amir Zadeh, Michael Chan, Paul~Pu Liang, Edmund Tong, and Louis-Philippe
  Morency. 2019.
\newblock Social-iq: A question answering benchmark for artificial social
  intelligence.
\newblock In \emph{Proceedings of the IEEE/CVF Conference on Computer Vision
  and Pattern Recognition}, pages 8807--8817.

\bibitem[{Zellers et~al.(2019)Zellers, Bisk, Farhadi, and Choi}]{vcr}
Rowan Zellers, Yonatan Bisk, Ali Farhadi, and Yejin Choi. 2019.
\newblock From recognition to cognition: Visual commonsense reasoning.
\newblock In \emph{Proceedings of the IEEE/CVF Conference on Computer Vision
  and Pattern Recognition}, pages 6720--6731.

\bibitem[{Zheng et~al.(2021)Zheng, Guha, Anderson, Henderson, and
  Ho}]{casehold}
Lucia Zheng, Neel Guha, Brandon~R Anderson, Peter Henderson, and Daniel~E Ho.
  2021.
\newblock When does pretraining help? assessing self-supervised learning for
  law and the casehold dataset.
\newblock \emph{arXiv preprint arXiv:2104.08671}.

\bibitem[{Zhu et~al.(2016)Zhu, Groth, Bernstein, and Fei-Fei}]{visual7w}
Yuke Zhu, Oliver Groth, Michael Bernstein, and Li~Fei-Fei. 2016.
\newblock Visual7w: Grounded question answering in images.
\newblock In \emph{Proceedings of the IEEE conference on computer vision and
  pattern recognition}, pages 4995--5004.

\end{thebibliography}
\bibliographystyle{acl_natbib}



\end{document}